\documentclass[letterpaper, 10 pt, conference]{ieeeconf}
\IEEEoverridecommandlockouts 
\let\labelindent\relax                                  
\usepackage{graphicx}
\usepackage{math}
\usepackage{amsmath}
\usepackage{color}
\usepackage{enumitem}
\usepackage{multirow}
\usepackage[linesnumbered,lined,ruled]{algorithm2e}

\begin{document}
\title{\LARGE \bf Aggregating Long-Term Context for Learning \\Laparoscopic and Robot-Assisted Surgical Workflows}
%\thanks{Supported by organization x.}
%\institute{}
\author{Yutong Ban$^{1,2}$,
Guy Rosman$^{1,3}$,
Thomas Ward$^{2}$, Daniel Hashimoto$^{2}$\\
Taisei Kondo$^{4}$, 
Hidekazu Iwaki$^{4}$, 
Ozanan Meireles$^{2}$, 
Daniela Rus$^{1}$
\thanks{$^{1}$ Computer Science and Artificial Intelligence Laboratory, 32 Vassar St, Cambridge, MA 02139, US. \{yban, rosman, rus\}@csail.mit.edu}
\thanks{$^{2}$ SAIIL, Department of Surgery, Massachusetts General Hospital. 55 Fruit Street, Boston, MA 02114, US. \{tmward, dahashimoto, ozmeireles\}@mgh.harvard.edu}
\thanks{$^{3}$ Toyota Research Institute, Cambridge, MA 02139, US.}
\thanks{$^{4}$ Olympus Corporation, Shinjuku Monolith, Shinjuku-ku, Tokyo, Japan. \{hidekazu.iwaki, taisei.kondo2\}@olympus.com}
}

\maketitle              % typeset the header of the contribution
\begin{abstract}
Analyzing surgical workflow is crucial for surgical assistance robots to understand surgeries. With the understanding of the complete surgical workflow, the robots are able to assist the surgeons in intra-operative events, such as by giving a warning when the surgeon is entering specific keys or high-risk phases. Deep learning techniques have recently been widely applied to recognizing surgical workflows. Many of the existing temporal neural network models are limited in their capability to handle long-term dependencies in the data, instead, relying upon the strong performance of the underlying per-frame visual models.
We propose a new temporal network structure that leverages task-specific network representation to collect long-term sufficient statistics that are propagated by a sufficient statistics model (SSM). We implement our approach within an LSTM backbone for the task of surgical phase recognition and explore several choices for propagated statistics. We demonstrate superior results over existing and novel state-of-the-art segmentation techniques on two laparoscopic cholecystectomy datasets: the publicly available Cholec80 dataset and MGH100, a novel dataset with more challenging and clinically meaningful segment labels.


\end{abstract}
%\begin{IEEEkeywords}
%\end{IEEEkeywords}

\section{Introduction}
\vspace{-2ex}
\label{sec:intro}
The future of robot-assisted laparoscopic surgery relies upon a strong automated understanding of surgical workflow from laparoscopic video. In order for robot systems to assist surgeons during surgery, they need a fundamental understanding of the surgical process. Surgical video represents an invaluable source of information as it is sufficient for surgical situational awareness and plentiful in the modern medical environment.
 While significant works has been performed for natural video analysis \cite{lin2020tsm, long2020purely, gan2015devnet, ban2016tracking, ban2019variational, xu2020train}, works also try to improve the understanding of video \cite{twinanda2016endonet,hashimoto2019surgical,kitaguchi_real-time_2019,zisimopoulos_deepphase:_2018,lalys2010surgical} and producing better annotation and supervision cues \cite{twinanda2016endonet,jin2020multi,qin2020temporal, WARD2020} for both laparoscopic and robot-assisted surgeries \cite{volkov2017machine},  existing models still fall short of a complete and automatic interpretation of surgery. A key cause of this performance gap is the manner in which surgery is interpreted from videos --- surgery is an inherently temporal process with a partially-observable state, and long-term temporal patterns. This is in contrast to other fields which have seen improved performance such as interpreting images from Computed Tomography (CT) or Magnetic Resonance Imaging (MRI) \cite{litjens2017survey}, where the patient state is more completely observed, or surgical technique analysis, where short-term interactions are sufficient.

%Computer vision (CV) excels in general object recognition~\cite{he2016deep}, object tracking~\cite{bertinetto2016fully} and image generation~\cite{goodfellow2014generative}. Works have  built-upon this foundation to apply deep learning models to the medical field, including automated disease diagnosis from Computed Tomography (CT) and Magnetic Resonance Imaging (MRI) scans \cite{litjens2017survey}. Even more recent applications have applied deep learning techniques to laparoscopic surgical videos. Surgical workflow recognition aims to recognize the different surgical steps in videos. Correctly identifying surgical steps is the foundation upon which to build computers' understanding of surgeries. However, accurate step recognition is challenging. First, surgical steps have fluid transitions which makes clear discrimination between step-transition difficult. Second, surgery workflow can vary significantly between surgeons and institutions, even for the exact same surgery. Third, organs and tissues in surgery are non-rigid and easily deform which makes them hard to analyze with current computer vision techniques. Last, identification of some surgery steps often needs reasoning across long time intervals (minutes to even hours), which makes the problem even more difficult for current temporal network model structure. 

Current computer vision efforts in surgical workflow analysis have
addressed operative phase recognition, as well as related tasks such as tool usage detection, tool segmentation and prediction of remaining surgery duration. In recent years, with the rise of deep learning methods, Convolutional Neural Networks (CNN)
with SVM and Hierarchical HMM being introduced in \cite{twinanda2016endonet} for surgical phase recognition. The
majority of approaches now use a CNN with a Long-Short Term Memory \cite{hochreiter1997long}
(LSTM) backbone, including \cite{hashimoto2019computer,aksamentov2017deep,twinanda2018rsdnet}. However, understanding surgical workflows requires reasoning about events across highly varied temporal scales, from a few seconds to hours, exceeding the capabilities of existing models. For example, in an endoscopic setting such as colonoscopy, identification of a polyp early in the procedure during insertion of the scope can influence the decision to perform polypectomy later during withdrawal. In laparoscopic cholecystectomy, ``Dissection of Calot's triangle'' involves removing the lower portion of the gallbladder from the liver bed (i.e. clearing the cystic plate). This phase can be visually indistinct from ``Removal of the Gallbladder from the Liver Bed'' later in the case and requires knowledge that key phases (occurring minutes later) have not yet occurred to accurately infer the current surgical phase. In such cases, information extracted by LSTM remains local compared to the total duration of the surgery and fails to improve classification performance of these cases.

% Initial works used combined model architectures
% of Support Vector Machines (SVM) with Hidden Markov Models (HMM)
% \cite{lalys2010surgical}.
% \cite{volkov2017machine} introduced an efficient model which uses coresets for segmenting the surgical workflows. By using coresets, the model can achieve identical levels of accuracy compared to training on entire visual cues while substantially reducing the training data to a small, but informative, subset.

Two main approaches have been recently proposed to ameliorate LSTM's shortcomings. Temporal and Dilated convolutional approaches take a multiscale approach and handle low-frequency processes well \cite{lea2016temporal}. Attention based models aimed at matching specific points in the distant past have been transformative for natural language processing and have been recently applied to surgical videos as well \cite{vaswani2017attention,shi2020lrtd}. Yet, neither of these approach capture more subtle phenomena such as aggregation of partial evidence from a segment in the distant past or order constraints between specific phases in the surgery that require measurements of interval lengths. While specific approaches have been attempted to cover specific priors on phase lengths and order \cite{jin2017sv, volkov2017machine}, a general framework that successfully handles long-term reasoning under uncertainty is still needed for surgical video understanding and remains an open problem.

In this paper, we address this problem by aggregating the long-term temporal context through sufficient statistic models (SSM) and combining them with visual cues to fed into an LSTM. 
% The SSM models then use the updated LSTM results to refine the global context features. 
The contribution of the paper is summarized as follows: 
\begin{enumerate}[label=(\roman*)]
    \item A novel SSM-LSTM framework that aggregates the long-term temporal context of different time scales to augment LSTM inference.
    \item An exploration of the various SSM feature choices and evaluation of their contributions.
    \item Validation of the proposed model on two large laparoscopic video datasets -- the publicly available Cholec80, and MGH100, a novel large-scale laparoscopic cholecystectomy video dataset. The proposed model obtains superior performance compared to state of the art methods on both datasets, with an advantage in clinically meaningful phenomena. 
\end{enumerate}

\section{Related works}
Surgical video analysis addresses several applications, such as phase recognition of laparoscopic surgery \cite{hashimoto2019computer,aksamentov2017deep,twinanda2018rsdnet} and  estimation of remaining surgery duration \cite{twinanda2018rsdnet}. An additional line of work explores videos from robotic-assisted surgery \cite{zhao2019real} and integrates kinematics and robotic system events
\cite{qin2020temporal,qin2020davincinet}. Significant efforts have been made on surgical instrument analysis \cite{saint2011detection,twinanda2016endonet,garcia2017toolnet}.  \cite{jin2019incorporating} integrate a priori information derived from motion flow into a temporal attention pyramid network for automatic instrument segmentation. \cite{zhao2020learning} extended this model by using a dual motion based semi-supervised framework, which leverages the self-supervised sequential cues in surgical videos. 
\cite{rivoir2020rethinking} uses a Bayesian network for anticipating the use of surgical instruments for context-aware assistance. Progress in the field is limited by data availability, which requires experts in surgery to label the data. \cite{tsai2019transfer} tries to overcome the problem by using transfer learning, which prevents the need to re-learn how to segment similar sub-tasks.
% learns motions’ traits from pre-segmented data, It is able to transfer the learned features and hyper parameters for new segmentation tasks, .

CNN-LSTM has become a standard architecture choice for processing video sequences with \cite{hashimoto2019computer,aksamentov2017deep,twinanda2018rsdnet} focusing on laparoscopic surgeries for phase recognition, and \cite{twinanda2018rsdnet} estimating a surgery's remaining time by adding a regression head. Moreover, \cite{kannan2019future} used CNN-LSTM jointly to predict the future state and recognizing different surgery types. Meanwhile, \cite{zisimopoulos2018deepphase} applied the architecture on cataract surgeries, and \cite{ward2020automated} used it for Per-Oral Endoscopic Myotomy (POEM) surgeries. Some variants of CNN-LSTM with modifications are proposed such as Prior
Knowledge Inference (PKI) (hand-crafted knowledge of operative phase
workflow to limit incorrect inferences of already finished phases)
\cite{jin2017sv} or "multitask" architectures that incorporate multiple streams
of inference in addition to operative phase labels to improve identification (e.g. tool
prediction \cite{jin2020multi}). 

In many of the methods above, the underlying temporal model is limited in its ability to analyze temporal information across time scales. HMMs are Markovian, while LSTMs are limited by their ability to propagate gradients in time \cite{bengio1994learning,pascanu2013difficulty,vaswani2017attention}. Models such as dilated temporal convolutions and hidden semi-Markov models (HSMMs) are limited in their ability to efficiently train information flow across long periods of time, with the latter limited by its inference computational efficiency. \cite{lea2016temporal} used a temporal convolution network (TCN) for action segmentation. \cite{liu2018deep} applied TCN in surgery and combined it with reinforcement learning (RL) for surgical gesture recognition. \cite{gao2020automatic} improved upon this using an uncertainty-aware tree search. 

Recently, several approaches tried to incorporate long-term temporal information by expanding the receptive field of traditional architectures. \cite{shi2020lrtd} used a non-local operation block on top of a CNN-LSTM framework to capture the long-range temporal dependencies. It acts similarly to a temporal attention network, to compute the similarity between the features of past and current frames. The past features are then linked to the current frame by using skip connections. \cite{czempiel2020tecno} proposed a Multi-Stage Temporal Convolutional Network (MS-TCN) that performs hierarchical prediction refinement for surgical phase recognition. The dilated convolutions used in the model allow for a larger receptive field. Instead of increasing the receptive field, our approach summarizes the global information from the surgery, as we aggregate main statistical features from the beginning of the surgery, as a side channel.

\begin{figure}[t!]
\centering
\includegraphics[width=.45\textwidth]{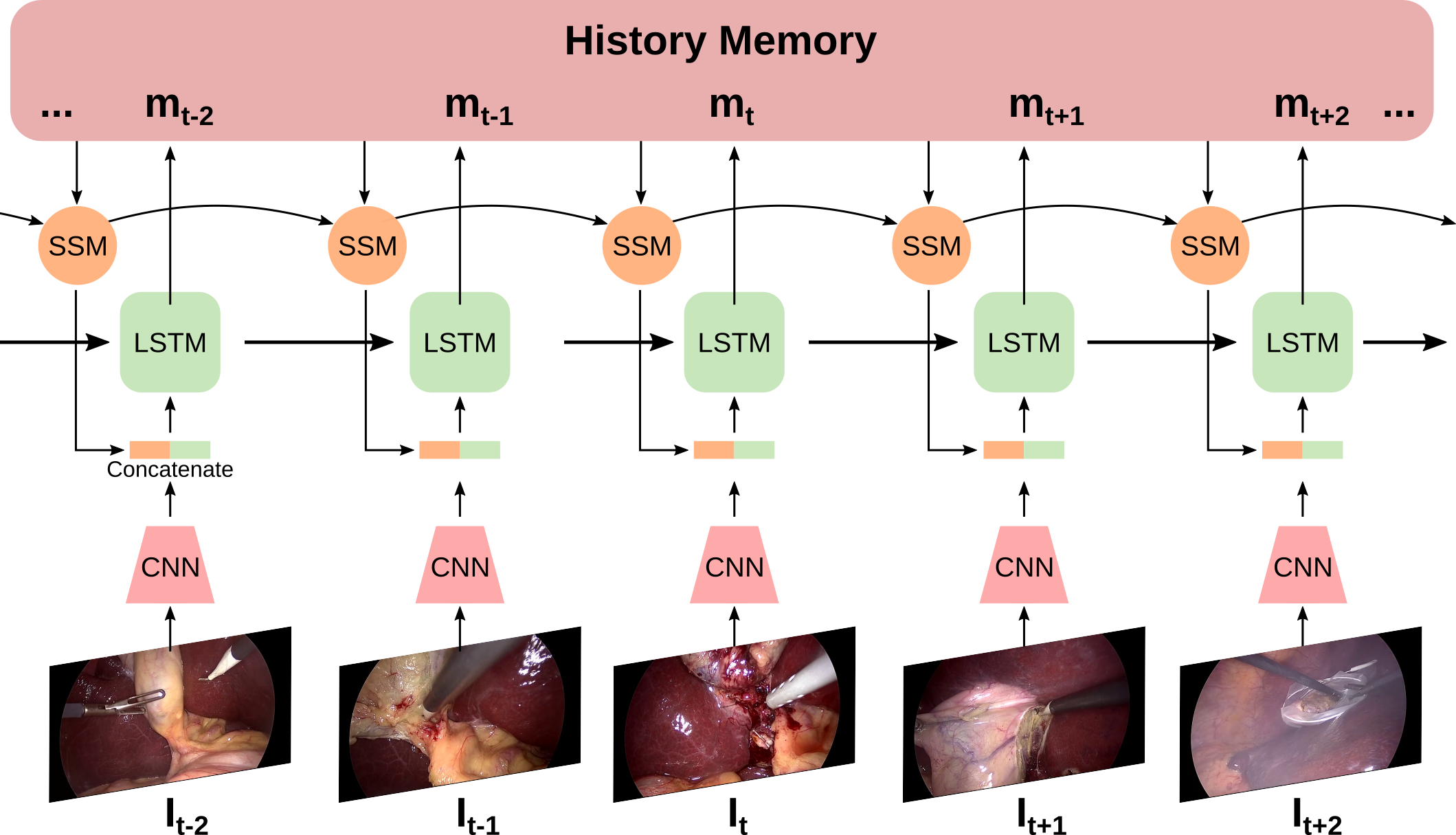}
\caption{SSM network architecture. Information from the network phase estimation head is processed as a multi-channel temporal signal. The resulting statistics are concatenated with the visual embedding and passed to the LSTM.}
\label{fig:overview}
\vspace{-2ex}
\end{figure}

\section{Methods}
We introduce an architecture to better leverage long-term temporal information in surgical phase recognition via approximate sufficient statistic features. We then proceed to detail a set of approximate sufficient statistic features included within the proposed architecture. 

\subsection{Model Architecture}
Surgical phase recognition attempts to classify the correct surgical phase label given video frames $\Imat_{t},t=0 \ldots T$.
% Suppose a video in our dataset is of length $T$ frames. 
% The video frames can be segmented into $N$ different surgical steps.
% Given the input surgery video frame  at time step $t$, the goal is to infer the correct step label for each frame, 
We denote the ground truth label for frame $\Imat_{t}$ by $y_{t} \in 1 \dots N$, where $N$ is the number of different surgical phases. 
We process individual frames via a CNN visual model (based on a ResNet module\cite{he2016deep}), encoding the visual content as a single vector $\vvect_{t}$, which is then fed to the LSTM, forming a standard CNN-LSTM structure. 

In analysis of temporal processes, recurrent neural networks such as LSTMs perform well when inference relies mainly on recent information. However, performance suffers when long-term temporal information is required for inference. To address the lack of long-term information, dilated convolutions \cite{DBLP:journals/corr/YuK15} have been suggested, but they fail to leverage several phenomena involved in the interpretation of surgeries:
\begin{enumerate}
    \item Correct classification of phase transitions depends on propagation of low-dimensional information that coincides with the actual phases being detected.
    \item Short events that are from the distant past can significantly affect interpretation of the current observations. (e.g. clearing the cystic plate is a visually identical task necessary in both ``Dissection of Calot's Triangle'' and ``Removal of Gallbladder from Liver Bed'' phases. The correct phase is identified from prior knowledge that the cystic structures have already been clipped and divided).
    \item Some of the temporal evidence collection occurs over a long period of time (consider priors about the length of each phase).
\end{enumerate}

While extracting a perfect sufficient statistic of the past is hard due to estimator dimensionality and the need to capture uncertainty about phase and perception limitations, the above phenomena make it possible to define a family of approximate sufficient statistics that can be computed from the data based on temporal aggregation of some transformation of the LSTM hidden state $L_t$.
This makes it easier for the network to do both short-range temporal reasoning (such as change detection and visual processing), as well as medium- and long-range reasoning (such as counting past frames of each phase).
The overall approach is presented in Algorithm~\ref{alg:forward_ssm} and illustrated in Figure \ref{fig:overview}.

\begin{algorithm}[t]
 \SetKwInOut{Input}{input}\SetKwInOut{Output}{output}
 \Input{Past LSTM hidden states $L_0\cdot L_t$, past sufficient statistic $s_0,\cdots,s_t$, next frame $I_{t+1}$ }
 \Output{next frame hidden state $L_{t+1}$, next frame sufficient statistic $s_{t+1}$, next frame phase estimate.}
 Compute visual model output $f_{V}(I_{t+1}$)\;
 Compute LSTM updated state $L_{t+1}$ given $f_{V}(I_{t+1}),s_t,L_t$\;
 Compute current phase estimate $y_{t+1}$ from $L_{t+1}$\;
 Compute frame statistics $s_F(L_{t+1})$\;
 Compute new sufficient statistics $s_{t+1}$ given $s_F(L_{t+1})$ and $s_{t}$.
 \caption{Forward estimate of surgery phase $y_{t+1}$.}
 \label{alg:forward_ssm}
\end{algorithm}
Our architecture takes the past hidden LSTM layer, and passes it through a transform (the phase recognition module), to get a vector $m_{t}$, conceptualized as a temporal vector signal $\Mmat _{t} = \{\mvect_{1} \ldots \mvect_{t} \}$. It then computes aggregate statistics of the transformed signal, resulting in a sufficient statistics feature stream $\Smat = \{\svect_{1}\ldots\svect_{t} \}$. By concatenating $\vvect_{t}$, it then feeds them to the current time phase LSTM inference as an augmented feature $\cvect_{t}$.
% A classic CNN-LSTM model would take $\vvect_{t}$ as the input to LSTM, then output the probability likelihood of each step $m_{tn}$. The proposed architecture extends the classic CNN-LSTM architecture: we first collect all the transformed LSTM states from the beginning of the video sequence and save them in a history memory $\Mmat = \{\mvect_{1}\dots\mvect_{t}\dots\mvect_{T} \}$. 
% Then different statistic models are applied to summarize the past information, resulting in a sufficient statistics feature stream $\Smat = \{\svect_{1}\ldots\svect_{t} \}$. By concatenating $\vvect_{t}$  with sufficient statistics feature vector $\svect_{t}$, an augmented feature $\cvect_{t}$ is obtained. 
% $\svect_{t}$ contains the long-term temporal information propagated from the start of the each video sequence.  
After concatenation, an LSTM is applied taking $\cvect_{t}$ as input to output the likelihood for each phase. Note that for both training and testing, the history memory $\Mmat_{t}$ is initialized with zeros. 

We note that several existing models fall within the family of functions described by this model, including temporal causal convolution (TCN) networks, PKI\cite{jin2017sv}, and LSTMs. Furthermore, several novel sufficient statistic features are detailed in Section \ref{subsec:ssm}. The LSTM output space captures an approximate sufficient statistic on the past. The SSM module extracts from past information a reduced set of approximate sufficient statistics to make inference in the current time-point more efficient.
Tailoring the choice of sufficient statistics can make it much easier for the network to learn specific dependencies and cues. While we describe this family regardless of resource constraints, in practice, many of the SSM features can be calculated in either $O(1)$ for computing-only or $O(1)$ for both computing and memory, as we will demonstrate.

\begin{table*}[ht]
\centering
\resizebox{1.0\textwidth}{!}{
\begin{tabular}{c|c|c|c}
\hline
 &  
\multicolumn{2}{|c|}{ \textbf{MGH100} } & \textbf{Cholec80}\\
\hline
Index & Phase Name & Description & Phase Name\\
\hline
0 & Port placement & Placement of ports for tool access to abdominal cavity &  Preparation\\
1 & Fundus retraction & Retraction of the GB fundus in preparation for dissection &  Calot Triangle Dissection\\
2 & Release GB peritoneum & Dissection of peritoneal lining of GB infundibulum& Clipping and Cutting\\
3 & Dissection of Calot's triangle & Dissection of the hepatocystic triangle to expose the cystic duct \& artery &Gallbladder Dissection\\
4 & Checkpoint 1 & Inactive period prior to the first clipping of a structure & Gallbladder Packaging\\
5 & Clip Cystic Artery & Application of surgical clips to the cystic artery & Cleaning and Coagulation\\
6 & Divide Cystic Artery & Division of the cystic artery & Gallbladder Retraction\\
7 & Clip Cystic Duct & Application of surgical clips to the cystic duct &-\\
8 & Divide Cystic Duct & Division of the cystic duct&-\\
% 9 & Checkpoint 2 & Inactive period between division of last structure in Calot's Triangle and the start of GB removal&-\\
9 & Checkpoint 2 & Inactive period between the end of the division \& the start of GB removal&-\\
10 & Remove GB from liver bed & Dissection of the GB from the liver &-\\
11 & Bagging & Placing the GB in a laparoscopic retrieval bag&-\\
12& Other step & Any other undefined step (free text annotation)&-\\
\hline
\end{tabular}
}
\caption{ Left: Phases in MGH100 and corresponding phase descriptions Right: Phases in Cholec80}
\label{tab:label_mgh_camma}
\label{tab:steps}
\vspace{-7ex}
\end{table*}
\subsection{Sufficient Statistic Features}
\label{subsec:ssm}

Different choices of summarization $S$ can make it easy for the network to learn long-term interactions. A few of the approaches explored in our experiments are described below.

%atisfying projection of the LSTM latent space to give good enough representation of temporal information. Therefore all the sufficient statistic features are built on the LSTM output space. Since the idea is to use the easy way of to encode the temporal information, in this paper, we propose three explainable sufficient statistic features.

\subsubsection{Hidden Markov Model (HMM)}
HMM is a well-known statistical method for temporal filtering of discrete states. It is thus intuitive to use HMM as a feature to encode temporal information. There are two main advantages of using HMM: 1) it provides smoother inference results, which provides an additional timescale of reasoning; 2) it can filter out impossible phase transitions using a state transition matrix, discouraging illogical phase transition inferences. 
\subsubsection{Cumulative Sum Likelihood (CSL)}
Temporal information can also be propagated by accumulating the LSTM inference likelihood in time:
\begin{equation}
\fvect_{t} = \log(\sum_{t' = 1}^t(\mathcal{I}_{l}(\mvect_{t'})) + 1),
\end{equation}
where $\fvect_{t}$ represents the CSL at time step $t$, and $\mathcal{I}_{l}$ represents thresholding of the elements of $\mvect$ with a set of threshold levels $l$, with respect to the maximum probability phase at time $t$. The CSL feature enhances the network's understanding of some global contexts and allows the network to capture both maximum-probability and probable interpretation of the phase at time $t$. It should have the capability to answer the question ``where we are'' in a surgery, including if certain phases have or have not already occurred. e.g. CSL can indicate that the phase ``Divide Cystic Duct''   
has already been achieved; since it is a non-repeated event, we know that future frames cannot be classified as such.

\subsubsection{Wavelets Transform} 
To capture temporal events at various time scales, we used a wavelet transform to summarize temporal information. 
% Different wavelet transformations can bring summaries of the temporal information in different aspects. 
We chose Gabor filter as a standard wavelet decomposition \cite{mallat1999wavelet}
which is defined as the modulation product of a Gaussian envelope and a complex sinusoidal wave, where the 1-dimensional Gabor filter can be formally written as:
% \begin{equation}
% Gabor(x) = \frac{1}{2 \pi \sigma} e^{-\pi \left[\frac{(x - x_{0})^2}{\sigma^{2}}\right]} e^{-i\xi x}    
% \end{equation}
% where $x_{0}$ is the center of the filter's spatial receptive field, $\sigma$ is the the standard deviations of the Gaussian envelope, and $\xi$ is the optimal spatial frequency of the filter in the frequency domain. 
We collect a filter bank with Gabor filters of different Gaussian envelope sizes to directly apply to the likelihood space along the time axis. The filtered results are then concatenated to gather the temporal information of different time scales.
% While this representation is $O(T)$ compute as described, there are efficient approximations for both Gabor and other wavelets. For example, Haar wavelets are trivial to compute at $O(1)$ complexity using integral images \cite{crow1984summed,Lewis95Fast}.

\subsubsection{Causal vs. acausal features} As the information can be propagated through time both in forward and backward directions, each of the features above can be built either in a causal manner or acausal manner, leveraging information from the future signal. We show the acausal SSM features as a proof of concept for offline analysis purposes \cite{twinanda2016endonet}. For certain phases (e.g. checkpoint 1 in MGH100 dataset), the inference of such phases may benefit from the future information.

\vspace{-3ex}
\section{Experiments}
\begin{figure}[h]
\vspace{-3ex}
\centering
\begin{tabular}{cc}
\includegraphics[width=0.24\textwidth]{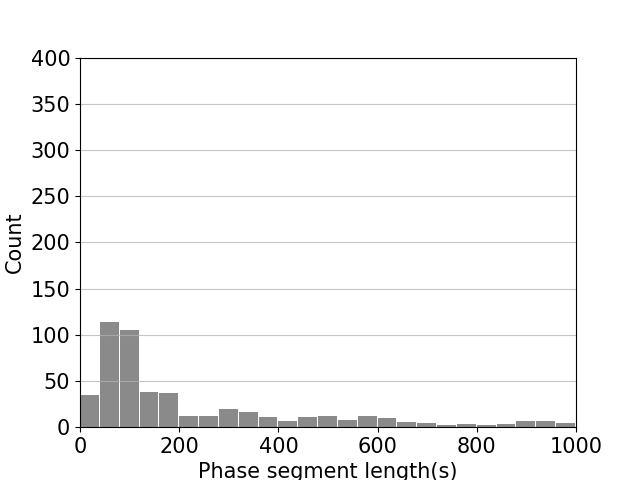}&
\hspace{-4ex}\includegraphics[width=0.24\textwidth]{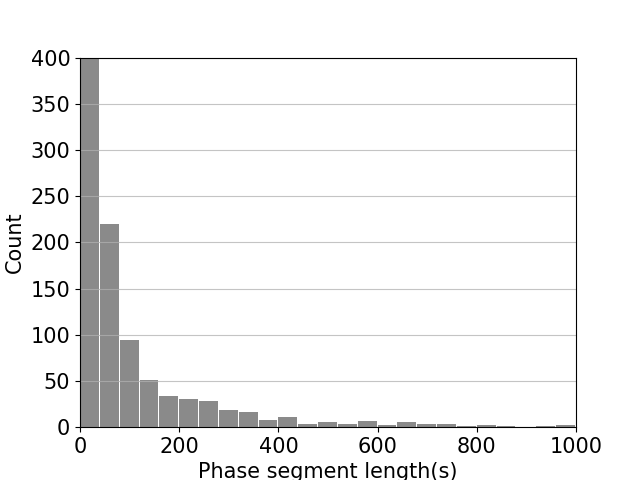}\\
(a) Cholec80 & (b) MGH100
\end{tabular}
\caption{Phase segment length distribution in the different datasets used in our experiments (x axis in seconds). Our dataset with phases as shown in Table 1 has a much larger distribution of small segments due to fragmentation of the phases of the surgery.}
\label{fig:results_mgh_length_distribution}
\vspace{-2ex}
\end{figure}
In this section, we introduce the details of the experimental setups, in addition to the results and discussions.
\subsubsection{Datasets}
We evaluated our method on two datasets:
\begin{enumerate}
    \item[] {\textbf{Cholec80}}  dataset\cite{twinanda2016endonet} contains 80 cholecystectomy videos each with 25 frames per second (fps). It provides annotations of tool presence and surgical phases (Table \ref{tab:label_mgh_camma}). The dataset is divided into a 40-40 split for training and testing. 
    \item[] {\textbf{MGH100}} dataset contains 100 cholecystectomy videos each with sampling rate of 30 fps. The phases in MGH100 (see Table \ref{tab:label_mgh_camma}) are more granular and clinically meaningful compared to Cholec80 (e.g., separating the broad category of ``clipping and cutting'' into separate tasks of clipping/cutting specific structures). We also added Checkpoints 1 and 2 to capture the decision points prior to clipping and removal of gallbladder, respectively.  80 videos are used for training while 20 are used for testing. 
\end{enumerate}

Prior work in surgical phase recognition has largely utilized public datasets such as Cholec80, which is annotated into long, visually distinct phases with linear progression. Clinically-meaningful phases, however, are often visually indistinct, of variable length, non-linearly progressing, repeating, and may be influenced by prior phases over the short and long-term -- such characteristics are reflected in the annotation structure of phases in MGH100 (Table \ref{tab:label_mgh_camma}). These phases can be considered clinically actionable (i.e. phases at which influencing a surgeon’s actions could modify risk of complications) and align more closely with surgical decision-making \cite{hashimoto2019surgical}.

\subsubsection{Evaluation Metrics}
We use standard metrics to evaluate our algorithms' overall performance. The phased-averaged recall and precision, and F1-score across phases, in addition total video per-frame accuracy (frames correctly inferred/total frames) \cite{padoy2012statistical} are used. 
\begin{figure*}[h!]
\centering
\includegraphics[width=0.75\textwidth]{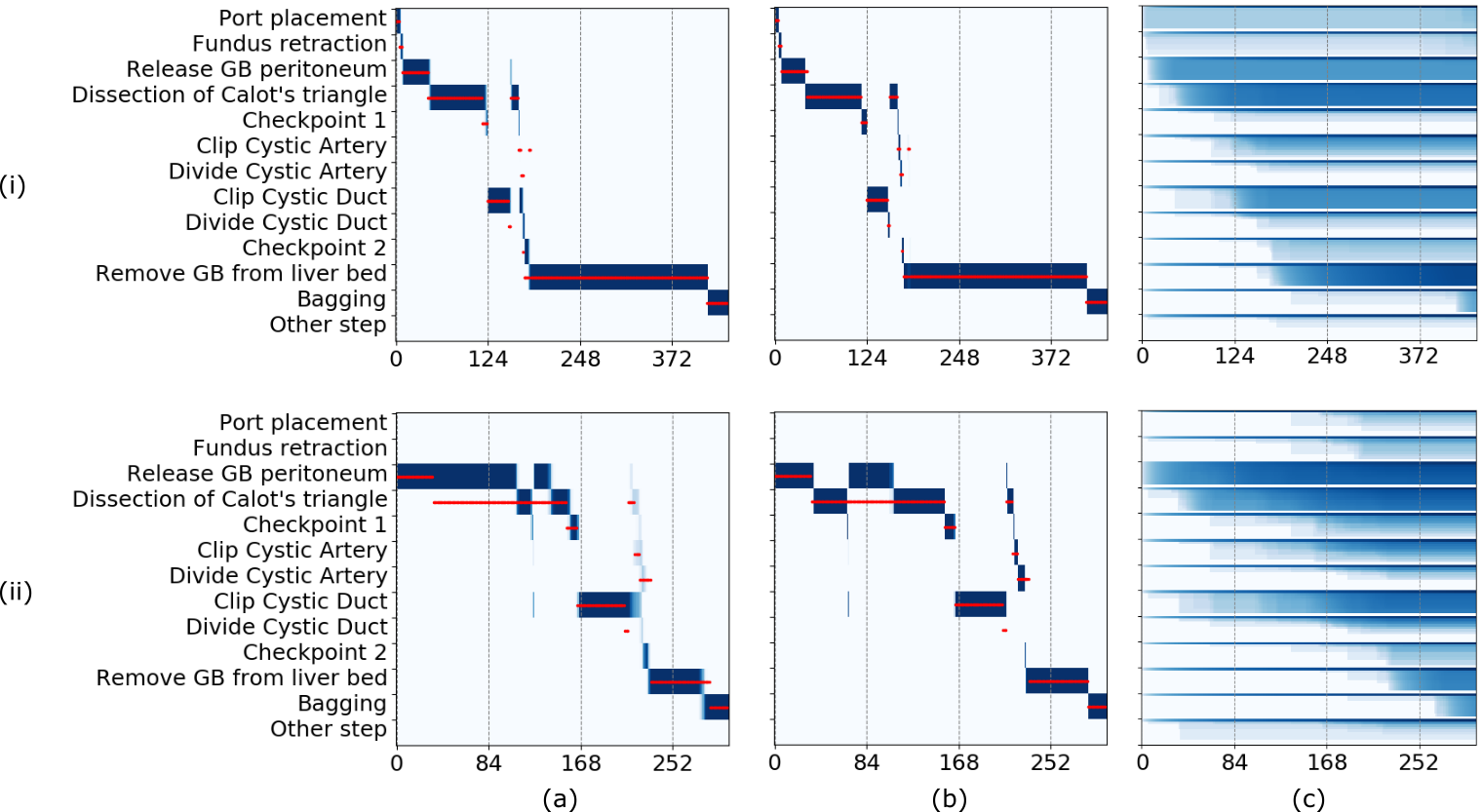}
% \begin{tabular}{cc}
% \includegraphics[width=0.4\textwidth]{figs/video_6_comparison.png}& \hspace{4ex}
% \includegraphics[width=0.36\textwidth]{figs/video_16_comparison.png}\\
% \end{tabular}
%\includegraphics[width=.7\textwidth]{figs/video6.png}
\caption{Example result in MGH100 dataset (i): video 6, (ii): video 17. The x-axis is the time in seconds (s); y-axis rows represent a phase. The red line indicates the Ground truth trajectories. (a) LSTM (b) SSM-LSTM. The blue bar indicates the estimation of the algorithm, with saturation proportional to certainty. (c) Cumulative  Sum  Likelihood (CSL) SSM features. The saturation of the blue bar indicates the accumulated values. In video 6, thanks to the SSM features, SSM-LSTM is able to accurately detect 'Clip/Divide Cystic Artery/Duct'. At the moment of 'Clip Cystic Artery', from (c) CSL feature the, 'Clip Cystic Duct' has already accomplished'. The current phase is therefore high likely to be 'Clip Cystic Artery'. Similar phenomenons in (ii).}. 
\label{fig:results_mgh_examples}
\vspace{-3ex}
\end{figure*}

\subsubsection{Model Parameters}
We down-sampled all the videos to 1 fps. We kept parameters identical for the SSM-LSTM and  LSTM for a fair comparison. The LSTM hidden state was 64-dimensions. During training, batch size was set to 32. An Adam \cite{kingma2014adam} optimizer was applied with a learning rate of 0.0025. The number of training epochs was 20. We first trained the CNN model for 20 epochs then fine-tuned the CNN model during temporal model training. For Gabor feature calculation, 10 different scales, $\sigma$, ranging from 10 to 30 frames, were applied. After that, features of different sizes were concatenated together.  
\subsubsection{Training Strategy}
To facilitate memory efficiency during training, we detached the computation of the sufficient statistics from its effect on recent past analysis via the LSTM, which was trained with temporal sequences of length 8. We numerically stabilized convergence via a lagged-step iteration, encouraging outputs to close to the previous epoch \cite{attouch2010proximal}.
% \begin{align}
% \nonumber
%     \theta^i =&\  \argmin_{\theta^i}\left\{\begin{array}{l}\|f_{\theta^i}(x,O^{i-1})-y\|^2+\\\alpha \|f_{\theta^i}(x,O^{i-1})-f_{\theta^{i-1}}(x,O^{i-1})\|^2\end{array}\right\}\\ 
%     O^{i-1} =&\  O(f_{\theta^{i-1}}(x,O^{i-2}))
% \end{align}
% \begin{figure}[t]
% \centering
% \includegraphics[width=0.5\textwidth]{figs/convergence.png}
% \caption{The log of the percentage of the residual in ssm features}
% \label{fig:results_mgh}
% \end{figure}
\subsubsection{Results}

The experiment results are shown both quantitatively and qualitatively. The results for Cholec80 dataset are in Table \ref{tab:quantitative results camma}, benchmarking the SSM-LSTM model against several state-of-the-art models. Despite only leveraging phase labels without incorporation of other features (tool, kinematics) like  \textit{EndoNet} \cite{twinanda2016endonet} or \textit{MTRCNet-CL} \cite{jin2020multi}, our proposed SSM-LSTM model has the best accuracy of 90.0\% among the different models, with similar performance across precision, recall and F1 score. The result of our proposed method is then followed by an HMM for further smoothing \cite{twinanda2016endonet}. The results of combining causal and acausal features are shown as a demonstration of offline applications, to gauge the effect of acausal information.
%, similar to \cite{twinanda2016endonet}. 
%By adding the acausal features, the model can achieve the accuracy of 90.8\%.  
\begin{table*}[ht]
\centering
\resizebox{.85\textwidth}{!}{
\begin{tabular}{c|c|ccccc}
\hline
&Model & Tool detection & Accuracy & Precision & Recall & F1 Score \\
\hline
&Binary Tool\cite{jin2020multi} & $\bullet$ & 47.5 & 54.4 & 60.2 & 57.2\\
&EndoNet\cite{twinanda2016endonet}  & $\bullet$ & 81.7 & 73.7 & 79.6 & 76.5 \\
Online&SV-RCNet\cite{jin2017sv}  & - & 85.3& 80.7 & 83.5 & 82.1 \\
&MTRCNet-CL\cite{jin2020multi}  & $\bullet$ & 89.2 & 86.9 & 88.0 & 87.4\\

% \color{blue}
% SSM-LSTM(Causal)& No& 89.5 & 83.3 & 87.3 & 85.1\\
&SSM-LSTM (Proposed)& - & 90.0 & 87.0 & 83.0 & 84.9\\
\hline
%SSM-LSTM(Causal 10 updates) & No& 89.5 & 86.5 & 82.7 & 84.6\\
Offline &Causal + Acausal SSM-LSTM (Proposed) & - & 90.8 & 85.3 & 82.7& 84.0\\
\hline
\end{tabular}
}
\vspace{-1ex}
\caption{Model performance on Cholec80 dataset. $\bullet$ denotes joint training of tools and phases. Our method outperforms phase-only approaches (-), and reaches a slightly better accuracy than tool-based approaches, despite no usage of tool information.}
\vspace{-5ex}
\label{tab:quantitative results camma}
\end{table*}

The results on the MGH100 dataset are in Table \ref{tab:quantitative results mgh overall}. LSTM is a baseline of CNN-LSTM structure similar to \cite{jin2017sv}. An ablation study of the proposed model using individual SSM features (e.g. Gabor, CSL) is also shown. The model's accuracy and F1 score benefited from multiple SSM features. With a combination of different SSM features, denoted as ``SSM'', the proposed model achieves the best performance in all the four metrics, significantly improving on LSTM.

\begin{figure}[!]
\vspace{-5ex}
\centering
\includegraphics[width=0.45\textwidth]{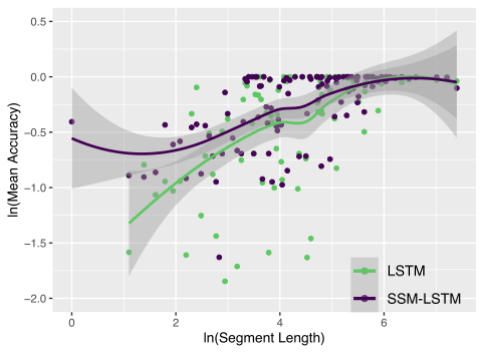}
\caption{Accuracy of phase segments in different length in MGH100 dataset. (plot in log space) Green: LSTM;
Purple, SSM-LSTM. Our approach consistently achieves better results compared to LSTM approach, especially for short and alternating annotation segments.}. 
\label{fig:results_mgh_curve}
\end{figure}

We also evaluated the performance of the model with phases of different lengths (Table~\ref{tab:quantitative results mgh phases}). We noticed the LSTM has a substantial performance drop in short phases (${<}30s$). However our SSM features helped ameliorate in such cases (cf. Accuracy). Our approach was able to improve significantly performance on short phases in this more challenging dataset, since the SSM provides additional information on the workflow structure. We also show the curve of the accuracy of the phases of different lengths in Fig \ref{fig:results_mgh_curve}, where we can see the SSM-LSTM (Purple) is significantly better than (LSTM) in short phases.

%  The trade-off between over-smoothing and evidence collection over multiple frames is fundamental in temporal process analysis.

\begin{table}
\centering
\resizebox{.5\textwidth}{!}{%
\begin{tabular}{c|c|cccc}

\hline
&Models & Accuracy & Precision  & Recall  & F1 Score  \\
\hline

&LSTM  & 83.3 & 50.2 & 51.6 & 50.9 \\
&Gabor & 84.1  & 51.2 & 57.4 & 54.1\\
Online&CSL & 85.4 & 58.8 & 60.0 &   59.4\\
&SSM & 85.6 & 59.4 & 61.5 & 60.4  \\
\hline
Offline&Acausal SSM &86.8 &61.9  &65.4 & 63.6\\
\hline
\end{tabular}
}
\caption{MGH100 dataset results. (SSM: the combination of all causal ssm features). Using SSM features, we improve both accuracy and F1 performance.}%
\label{tab:quantitative results mgh overall}
\vspace{-3ex}
\end{table}

 We qualitatively show how applying SSM features assists the model to understand the temporal structure of the surgery in Fig~\ref{fig:results_mgh_examples}. For both examples, the SSM-LSTM model can accurately detect the short phases ``Clip Cystic Artery'' and ``Clip Cystic Duct.''  which are more clinically meaningful phase labels, the ground truth labels included alternating patterns, which differ from traditional linear workflows \cite{twinanda2016endonet,volkov2017machine,kitaguchi_real-time_2019,hashimoto2019computer,zisimopoulos_deepphase:_2018}. They are two phases which are hard to distinguish based only on the visual model, since the same surgical tool (clip applier) is used. Moreover, artery and duct are spatially close to each other, making the problem even harder. We can see from Fig~\ref{fig:results_mgh_examples} (i-b)  that SSM-LSTM has a better performance. That is due to the fact when the first 'Clip Cystic Duct' has accomplished, the network gets the 'Duct has been clipped' information from the SSM feature. Therefore, when the model sees the clip applier appear the second time in the video, it is high likely the surgeon is going to ``Clip Cystic Artery''. Contrast this with the  LSTM, which, lacking long-term context, was not able to identify the two phases correctly. Similar phenomena are observed in video 17 in Fig~\ref{fig:results_mgh_examples} (ii).

\begin{figure}[!]
\centering
\begin{tabular}{c}
 \hspace{4ex} \includegraphics[width=.38\textwidth]{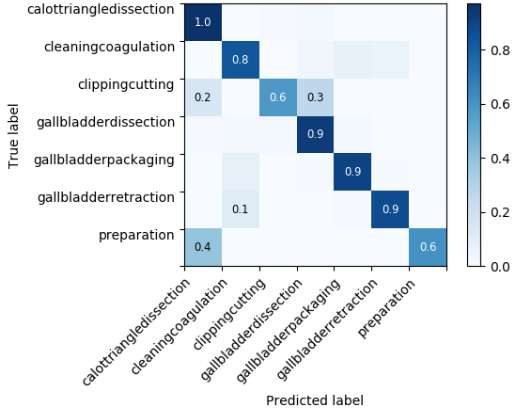}  \\ 
 \includegraphics[width=.4\textwidth]{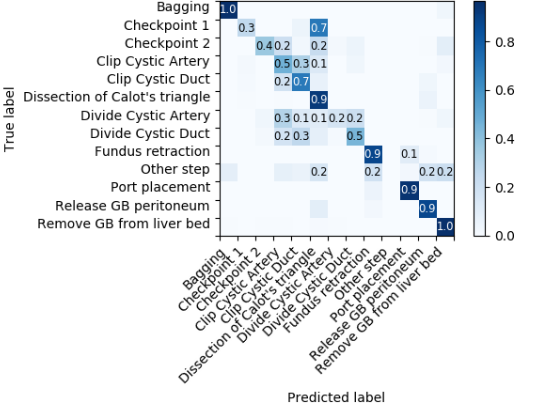}  
 \vspace{-2ex}
\end{tabular}
\caption{Confusion matrix for each phase on Cholec80 dataset  (up), MGH100 dataset (down). The proposed method has good performance on Cholec80 phases. On MGH100, the model performs well on long phases (e.g. Release GB Peritoneum), but for the phases which are short and cannot be assisted by temporal cues, the performance are not as good. e.g. Checkpoint 1.}. 
\label{fig:cm}
\end{figure}

We also analyzed SSM-LSTM performance on the individual phases, shown in Fig. \ref{fig:cm}. On the MGH100 dataset, the algorithm had good performance on long phases such as ``Release GB Peritoneum'' and ``Dissection of Calot’s Triangle'', with accuracy over $90\%$. Short phase performance was worse, as some of the short phases are likely harder to infer due to the lack of data variability (e.g. $37\%$ of accuracy for phase checkpoint 2). However, with SSM, short phase performance exceeded that of LSTM, shown in Table \ref{tab:quantitative results mgh phases}.

\begin{table}
\centering
\resizebox{.45\textwidth}{!}{%
\begin{tabular}{c|c|ccccccccc}

\hline
&Models & 1-3s & 4-10s & 11 -30s & 31 - 60s & $>$60s \\
\hline
%Classic LSTM  & 82.4 & 49.2 & 50.6 & 49.8 \\
&LSTM & 12.5 &  40.6 &  47.0 &  63.1  & 91.7\\
%\color{blue}
%HMM-LSTM &  80.04 &60.64 & 53.71  & 40.77 \\
%CSL-LSTM & 83.9 & 52.4 & 53.6 &   53.0 \\
&Gabor&  15.6 &  44.1&  53.4&  66.4 &  90.2\\
Online&CSL & 20.3 &  45.6  &  54.7&  63.5& 93.0\\
%Causal SSM-LSTM & 84.3 & 53.8 & 54.8 & 54.3 \\
&SSM & 31.2& 51.6& 58.2&  64.4&  93.0 \\
\hline
%Causal+ Acausal SSM-LSTM &85.0 &54.6  &57.3 & 55.9 \\
Offline&Acausal SSM &35.9 & 52.4 & 64.3& 64.7& 93.0\\
%Causal+ Acausal SSM-LSTM &87.7 &60.0  &59.6 & 59.8 \\
\hline
\end{tabular}
}
\caption{Accuracy for different phase durations. Our algorithm significantly improves short and challenging phase segments (duration $<$ 30s).}%
\label{tab:quantitative results mgh phases}
\vspace{-7ex}
\end{table}

In addition, we evaluated the accuracy of phase transitions and phase midpoints as per-segment statistics. For transition accuracy, a transition estimated within 10s of ground truth was considered correct. The SSM-LSTM achieved transition accuracy of $48.1\%$ vs. LSTM  at $39.0\%$. The accuracy for phase midpoint was $63.69\%$ for SSM-LSTM vs. $56.23\%$ for LSTM. The performance benefits from the SSM module for both phase and transition inference, likely due to a better inference of phase start/end points and phase duration.

\vspace{-0.1cm}
\section{Conclusions}
In this paper we propose a novel SSM-LSTM model that aggregates temporal information to augment LSTM inference. The proposed model is validated on two large surgery datasets, Cholec80 and MGH100, surpassing state-of-the-art performance on a more clinically relevant and harder taxonomy. We demonstrate the advantage of the proposed model over the existing methods in several clinically important scenarios. Using different SSM features, the model benefits from complementary approaches for temporal analysis, and improves understanding long-range, clinically relevant temporal interactions in surgical workflows. 

\bibliographystyle{IEEEtran}
\bibliography{root}

\end{document}